\documentclass{article}
\usepackage{spconf,amsmath,graphicx}

\title{FocusNet: An attention-based Fully Convolutional Network for Medical Image Segmentation}
\name{Chaitanya Kaul, Suresh Manandhar, Nick Pears\thanks{Accepted as an ISBI 2019 conference paper}}
\address{Department of Computer Science, University of York, Heslington, York, United Kingdom, YO10 5DD}
\begin{document}
%
\maketitle
\begin{abstract}
We propose a novel technique to incorporate attention within convolutional neural networks using feature maps generated by a separate convolutional autoencoder. Our attention architecture is well suited for incorporation with deep convolutional networks. We evaluate our model on benchmark segmentation datasets in skin cancer segmentation and lung lesion segmentation. Results show highly competitive performance when compared with U-Net and it's residual variant. 
\end{abstract}
\section{Introduction}
\label{sec:intro}

Convolutional Neural Networks (CNNs) have had great success in in various computer vision tasks \cite{vgg} \cite{unet} \cite{yolo}. These architectures eradicate the need for hand crafted features, leading to training end-to-end systems that are able to learn important features and simultaneously perform the required task leading to state-of-the-art accuracy. As a consequence, CNNs have quickly become the baseline for most computer vision tasks. Techniques such as dropout have been used as a means of regularization to increase the generalizability of deep neural networks. Batch normalization improves the vanishing gradient problem in deep networks. Despite these techniques, the accuracy tends to stagnate with very deep CNNs. Architectures such as ResNet \cite{resnet} address this issue by proposing residual learning that permits very deep CNNs without sacrificing accuracy. Identity mappings \cite{identity} has been proposed as an enhancement to the residual blocks used within ResNet. \\
More recently attention based methods have been developed that allow a network to focus on the most relevant parts of the data \cite{qa} \cite{lipread}. These techniques of attention have been applied to tasks such as visual question answering \cite{qa} and lip reading in the wild \cite{lipread}, but their application to refine prediction in medical imaging is still to be tested. Self attention blocks such as those implemented within SE nets \cite{se} re-calibrate the response of network filters using \textit{squeeze} and \textit{excitation} blocks.\\
In this paper, we propose a general architecture for combining a separate attention mechanism into a ResNet + SE network hybrid architecture. We demonstrate that our attention mechanism 
improves on an already state-of-the-art ResNet + SE architecture. For medical image processing tasks, existing works \cite{histeq} \cite{denoising} \cite{contrasteq} demonstrate that pre- and post-processing such as, histogram equalization \cite{histeq}, image denoising \cite{denoising} , contrast equalization \cite{contrasteq}, are crucial in achieving high accuracy. We demonstrate that our proposed architecture requires very minimal processing and performs well on images with varying conditions. \\
The paper is organized as follows: The next section describes our architecture. Section \ref{sec:eval} describes the two datasets used, along with our evaluations. Section \ref{sec:res} presents our results, and a final section is used for conclusions. 

\section{Attention Architecture}
\label{sec:arch}
\begin{figure*}[htb]
\begin{center}
\includegraphics[scale=0.34]{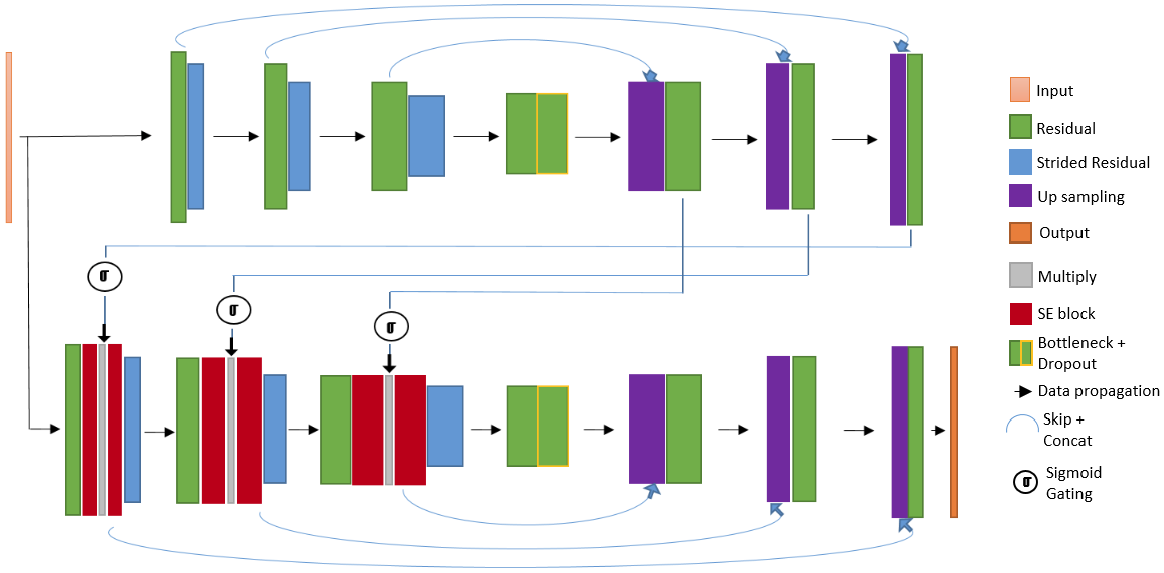}
\end{center}
\caption{Our network architecture uses attention to give better per pixel predictions, leading to better segmentation. The two branches are comprised of encoder-decoder structures where the per-layer decoded output is passed through a sigmoid gating function and multiplied with the output of the first SE block. The arrows show the direction of information flow in the network.}
\label{arch}
\end{figure*}

A conventional autoencoder first creates a low dimensional representation of the input and then upsamples from that representation to recreate the input. We exploit this encoder-decoder architecture to hierarchically extract latent attention maps leading to more accurate decoding. We propose the FocusNet architecture (see Fig.\ref{arch}), that employs two parallel branches of information flow with one branch solely devoted to attention. The attention branch employs an encoder-decoder structure with skip connections from the encoder to the decoder to facilitate better gradient flow. Our architecture provides a strong bias for the two networks to specialise and learn different representations.\\
Given an image, $x_i \in X$ where $X$ is the mini batch, each layer of an encoder learns a mapping $G$, given by,
\begin{equation}
    E_\textit{l} = G_\textit{l}(x, \mathbf{W_{\textit{l}}})
\end{equation}
The decoder corresponding to this layer, decodes this representation in the following form:
\begin{equation}
    D_\textit{l} = [G_\textit{l}(x, \mathbf{W_{\textit{l}}}) \ ; \ H_l(H_{l-1}(x, \mathbf{W_{\textit{l}-1}}))]
\end{equation}
where $[ \ ; \ ]$ denotes concatenation via skip connection and $H_{l-1}$ is the output from the previous decoder layer.\\ 
In the encoder in the second branch, the output can be represented as:
\begin{equation}
    A_\textit{l} = F_\textit{l}(x, \mathbf{W_{\textit{l}}}) \cdot \sigma(D_\textit{l})
\end{equation}
where $A_l$ is the output of the $l^{th}$ layer of the second encoder after \textit{gating-and-multiplication}. \\
We use no bottleneck full pre-activation residual blocks \cite{identity} in the second branch (see bottom of Fig.\ref{arch}). Downsampling is done using strided convolutions rather than max pooling. The output is a 1x1 convolution with sigmoid activation that outputs per-pixel predictions. The rest of the convolutions have a 3x3 receptive field. The filter bank volumes used are $32 \xrightarrow{} 64 \xrightarrow{} 128 \xrightarrow{} 256 \xrightarrow{} 512 \xrightarrow{} 256 \xrightarrow{} 128 \xrightarrow{} 64$. \\
Skip connections throughout the architecture facilitate better gradient flow, leading to easier training of a deeper network. Additionally, SE blocks are used extensively to recalibrate the weighting of the output feature maps at intermediate steps. For hyperparameter settings, we use dropout with a fixed rate of 0.2 throughout.

\begin{figure}[htb]
\begin{minipage}[b]{0.32\linewidth}
  \centering
  \centerline{\includegraphics[width=0.8\textwidth]{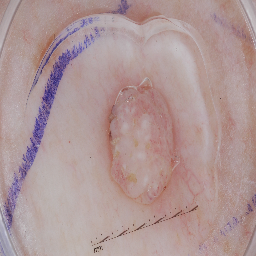}}
\end{minipage}
\begin{minipage}[b]{0.32\linewidth}
  \centering
  \centerline{\includegraphics[width=0.8\textwidth]{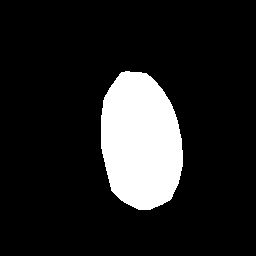}}
\end{minipage}
\hfill
\begin{minipage}[b]{0.32\linewidth}
  \centering
  \centerline{\includegraphics[width=0.8\textwidth]{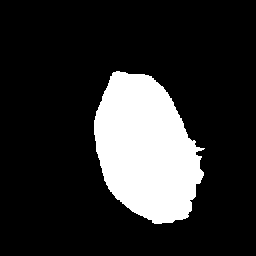}}
\end{minipage}
\begin{minipage}[b]{0.32\linewidth}
  \centering
  \centerline{\includegraphics[width=0.8\textwidth]{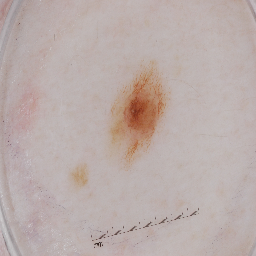}}
\end{minipage}
\begin{minipage}[b]{0.32\linewidth}
  \centering
  \centerline{\includegraphics[width=0.8\textwidth]{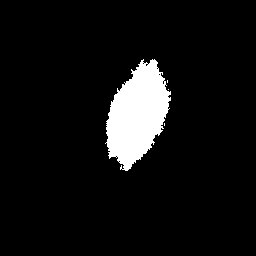}}
\end{minipage}
\hfill
\begin{minipage}[b]{0.32\linewidth}
  \centering
  \centerline{\includegraphics[width=0.8\textwidth]{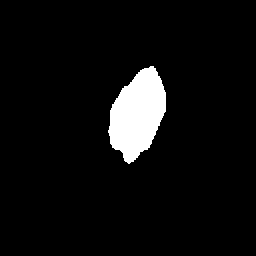}}
\end{minipage}
\begin{minipage}[b]{0.32\linewidth}
  \centering
  \centerline{\includegraphics[width=0.8\textwidth]{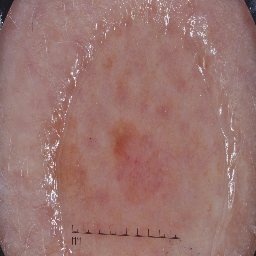}}
\end{minipage}
\begin{minipage}[b]{0.32\linewidth}
  \centering
  \centerline{\includegraphics[width=0.8\textwidth]{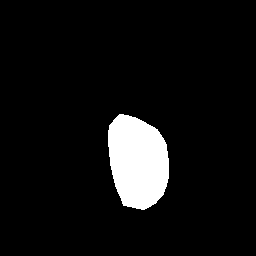}}
\end{minipage}
\hfill
\begin{minipage}[b]{0.32\linewidth}
  \centering
  \centerline{\includegraphics[width=0.8\textwidth]{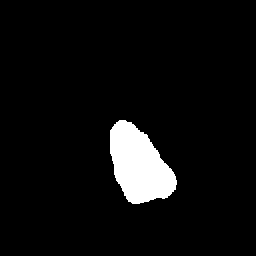}}
\end{minipage}
\begin{minipage}[b]{0.32\linewidth}
  \centering
  \centerline{\includegraphics[width=0.8\textwidth]{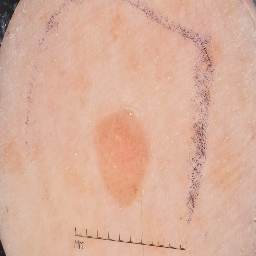}}
\end{minipage}
\begin{minipage}[b]{0.32\linewidth}
  \centering
  \centerline{\includegraphics[width=0.8\textwidth]{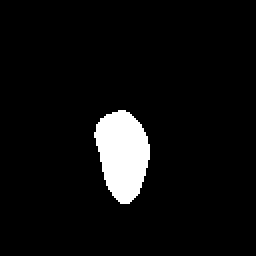}}
\end{minipage}
\hfill
\begin{minipage}[b]{0.32\linewidth}
  \centering
  \centerline{\includegraphics[width=0.8\textwidth]{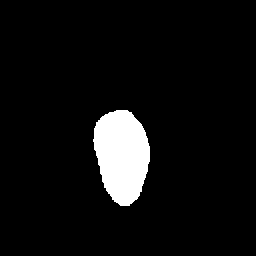}}
\end{minipage}
\caption{Experimental segmentation results on the melenoma dataset. Column 1 is the input image, column 2 is the ground truth, and column 3 is the segmentation. The algorithm achieves good results given the problem setup, but the data clearly requires some amount of pre-processing to get better accuracy.}
\label{fig:res}
\end{figure}




\section{Evaluation}
\label{sec:eval}

\subsection{Datasets}
To demonstrate the performance of our architecture, we evaluate it on two different medical imaging datasets: skin cancer lesion segmentation and lung segmentation, as follows.

\subsubsection{Skin Cancer Segmentation}
This dataset \cite{skincancer} is a part of the ISIC skin cancer segmentation challenge on finding regions of melanoma inside skin images. The dataset contains 2000 RGB images in its training set, 150 images in its validation set and 600 test images. The images are high resolution and of varying sizes. The masks are 8-bit grayscale images with intensity value zero representing the background, and intensity value 255 representing the cancer region. 

\subsubsection{Lung Segmentation}
The lung segmentation dataset \cite{lungseg} contains 2D images in .tif format with provided ground truth segmentation maps. The images are single channel with the size of 512x512 pixels. The dataset is fairly small, containing a total of 267 images.

\subsection{Experiment Details}
All experiments were done in python, using Keras \cite{keras} with a Tensorflow backend. Two Nvidia GTX 1080ti GPUs were used for all experiments. The batch size for training and testing was kept at 8. We trained all segmentation experiments using the dice coefficient loss. We subtracted the value of the loss from 1 to get a value between [0,1] for mathematical convenience, such that the loss could converge reducing towards zero. The Adam optimizer was used with the default learning rate. We reduced the learning rate on a plateau by half, if the validation loss didn't improve by an epsilon of 0.001 over 5 consecutive epochs. The validation loss was also monitored at every epoch such that the best performing model on the validation set could be saved. All experiments were run for a maximum of 80 epochs. \\
Images from both the datasets were resized to 256x256 pixels. No pre-processing was used other than subtracting the mean pixel value for the respective datasets and dividing by the standard deviation to normalize it. This was necessary so that the network wouldn't run into the exploding gradient problem. The lung segmentation data was divided into a 80-20 training-validation split. The skin cancer segmentation data came with its own validation set and was used as-is.\\
We used data augmentation to increase the size of our datasets. For the lung cancer dataset, we used random zooms and flipping to increase the training images size to 1700. For the skin cancer dataset, we also added a very small number of random channel shifts to generate an augmented dataset of 6000 training images. 

\subsection{Evaluation Metrics}
To evaluate the performance of our network, we used the following metrics. We measured the Dice (DI) and Jaccard index (JI) values for all networks as these are standard metrics in evaluating segmentation results. Along with them, we also measured the accuracy (AC), sensitivity (SE) i.e. the true positive rate and specificity (SP) i.e. true negative rate values for our results.

\section{Results}
\label{sec:res}
 Table 1 provides the results on the lung segmentation dataset. We compared our results with the recurrent version of the U-Net, and a recurrent U-Net architecture with residual mappings instead of convolutions. Compared to the R2U-Net \cite{recunet} and the corresponding recurrent class of architectures presented in that paper, our network outperforms in every metric except the sensitivity value.\\
 Table 2 gives the results on the test set of the skin cancer dataset for detecting melanoma. It can be seen that our results are comparable with the recent results on this dataset. Compared to the LIN architecture \cite{linmelenoma2}, it can be observed that even without any pre- and post-processing, our network outperforms the architecture (in terms of the Jaccard Index, which was the evaluation metric of the competition) due to how our network incorporates a bias in the second branch of the encoder, forcing it to learn more robust features from the data. The LIN architecture, on the other hand, relied heavy image pre-processing. \\
 
 \begin{table}[htb]
\label{results_melenoma}
\begin{center}
\begin{tabular}{|c||c|c|c|c|c|}
\hline
Method & SE & SP & AC & JI\\
\hline
\hline
U-Net \cite{recunet} & 0.9696 & 0.9872 & 0.9828 & 0.9858 \\
\hline
Res-U-Net \cite{recunet} & 0.9555 & 0.9945 & 0.9849 & 0.9850 \\
\hline
RU-Net \cite{recunet} & 0.9734 & 0.9866 & 0.9836 & 0.9836 \\
\hline
R2U-Net \cite{recunet} & 0.9826 & 0.9918 & 0.9897 & 0.9897 \\
\hline
R2U-Net \cite{recunet} & \textbf{0.9832} & 0.9944 & 0.9918 & 0.9918 \\
\hline
FocusNet (ours) & 0.9757 & \textbf{0.9981}  & \textbf{0.9932} & \textbf{0.9965} \\
\hline
\end{tabular}
\end{center}
\caption{Segmentation results on the validation set for lung segmentation dataset. We extend the table presented by \cite{recunet} with our results on the dataset.}
\end{table}

\begin{table*}[htb]
\label{results_melenoma}
\begin{center}
\begin{tabular}{|c||c|c|c|c|c|}
\hline
Method & SE & SP & AC & JI & DI\\
\hline
\hline
FCN-8s \cite{linmelenoma2} & 0.806 & 0.954 & 0.933 & 0.696 & 0.783\\
\hline
U-Net \cite{unet} & 0.853 & 0.957 & 0.920 & 0.651 & 0.768 \\
\hline
II-FCN \cite{IIfcn} & 0.841 & 0.984 & 0.929 & 0.699 & 0.794 \\
\hline
Auto-ED \cite{autoed} & 0.836 & 0.966 & \textbf{0.936} & 0.738 & 0.824 \\
\hline
Thao \textit{et al.} \cite{melanoma1} & 0.6513 & 0.9421 & 0.8772 & 0.5065 & 0.6317\\
\hline
LIN \cite{linmelenoma2} & \textbf{0.855} & 0.974 & 0.934 & 0.753 & \textbf{0.839} \\
\hline
FocusNet (ours) & 0.7673 & \textbf{0.9896} & 0.9214 & \textbf{0.7562} & 0.8315 \\
\hline
\end{tabular}
\end{center}
\caption{Segmentation results on the test set for skin cancer detection. We extend the table presented by \cite{linmelenoma2} with a few more results \cite{autoed}, \cite{melanoma1}, including ours. The results on the FCN and U-Net are reported from \cite{linmelenoma2} and have been trained on data pre-processed using their strategy.}
\end{table*}


\section{Conclusions}
In this paper, we presented our novel attention-based deep neural network architecture, FocusNet. The architecture learns a better encoding of the data in its encoder layers to predict robust segmentation maps, making it a useful architecture to employ in image segmentation tasks. The residual nature of the convolutions used, facilitates training deeper neural networks that don't overfit. The skip connections used throughout help in gradient flow through the entire network. A drawback of the network is its lack of responsiveness to sensitivity metric of the data. We believe it is partly due to the fact that we trained the architecture without applying any pre- or post-processing on the datasets. Our method of incorporating attention in CNNs can be easily generalized to other domains of computer vision in addition to segmentation. 


\bibliographystyle{IEEEbib}
\bibliography{strings,refs}

\end{document}